%% file: main.tex
\documentclass[10pt,twocolumn,letterpaper]{article}

\usepackage{cvpr}              % To produce the CAMERA-READY version
\usepackage{kotex}
\usepackage{amsmath}
\usepackage{algorithm}
\usepackage{algorithmic}
\usepackage{amsthm}

\usepackage{thm-restate}
\DeclareMathOperator*{\argmax}{arg\,max}
\usepackage{wrapfig}
\usepackage{bbm}
\usepackage{multirow}
\usepackage{tabularx}
\usepackage{tabulary}
\usepackage{siunitx}
\usepackage{subcaption}
\usepackage{caption}
\usepackage[utf8]{inputenc} % allow utf-8 input
\usepackage[T1]{fontenc}    % use 8-bit T1 fonts
\usepackage{url}            % simple URL typesetting
\usepackage{booktabs}       % professional-quality tables
\usepackage{amsfonts}       % blackboard math symbols
\usepackage{nicefrac}       % compact symbols for 1/2, etc.
\usepackage{microtype}      % microtypography
\usepackage{xcolor}         % colors


\definecolor{cvprblue}{rgb}{0.21,0.49,0.74}
\usepackage[pagebackref,breaklinks,colorlinks,allcolors=cvprblue]{hyperref}

\def\paperID{-} % *** Enter the Paper ID here
\def\confName{CVPR}
\def\confYear{2026}

\title{FALCON: False-Negative Aware Learning of Contrastive Negatives in Vision-Language Alignment}

\author{
    Myunsoo Kim\thanks{Equal contribution} \qquad
    Seongwoong Shim\footnotemark[1] \qquad
    Byung-Jun Lee\thanks{Corresponding author} \\
    Korea University \\
    {\tt\small \{m970326, ssw030830, byungjunlee\}@korea.ac.kr }
}

\begin{document}
\maketitle
\input{sec/0_abstract}
\input{sec/1_intro}

\input{sec/2_motivation}
\input{sec/3_main}
\input{sec/4_exp}
\input{sec/5_conclusion}
{
    \small
    \bibliographystyle{ieeenat_fullname}
    \bibliography{main}
}

\input{sec/X_suppl}

\end{document}

%% file: sec/0_abstract.tex
\begin{abstract}
False negatives pose a critical challenge in vision-language pretraining (VLP) due to the many-to-many correspondence between images and texts in large-scale datasets. These false negatives introduce conflicting supervision signals that degrade the learned embedding space and diminish the effectiveness of hard negative sampling. In this paper, we propose FALCON (False-negative Aware Learning of COntrastive Negatives), a learning-based mini-batch construction strategy that adaptively balances the trade-off between hard and false negatives during VLP. Rather than relying on fixed heuristics, FALCON employs a negative mining scheduler that dynamically selects negative samples of appropriate hardness for each anchor instance during mini-batch construction, guided by a proxy for cross-modal alignment improvement. Experimental results demonstrate that FALCON significantly improves performance across three vision-language learning frameworks (ALBEF, BLIP-2, SigLIP-2) and a broad range of downstream tasks and evaluation settings, underscoring its effectiveness and robustness in mitigating the impact of false negatives.
\end{abstract}

%% file: sec/1_intro.tex
\section{Introduction}
\label{introduction}
The goal of Vision-and-Language Pretraining (VLP) is to learn cross-modal representations from large-scale image-text pairs, improving performance on downstream tasks such as image-text retrieval (IRTR) \citep{frome2013devise}, visual question answering (VQA) \citep{antol2015vqa}, and natural language for visual reasoning (NLVR) \citep{suhr2018corpus}. Recent advancements \citep{li2021align, li2022blip, li2023blip} have demonstrated remarkable progress in this domain, underscoring the effectiveness of VLP in cross-modal representation learning. 

These models are typically trained with self-supervised objectives such as masked language modeling (MLM), image-text contrastive (ITC), and image-text matching (ITM) losses. While ITC and ITM are effective in improving the quality of the learned embedding space, they inherently require negative samples during training. In particular, the inclusion of \textit{hard negatives}---those that are semantically similar to the anchor sample---has been shown to be crucial for the success of VLP~\citep{schroff2015facenet}. As a result, substantial research efforts have been devoted to devising more effective strategies for choosing such hard negatives~\citep{li2021align,li2022blip,li2023blip,byun2022grit}.

\begin{figure}[t!]
    \centering
    % \vspace{1.0cm}
    \includegraphics[width=0.9\linewidth]{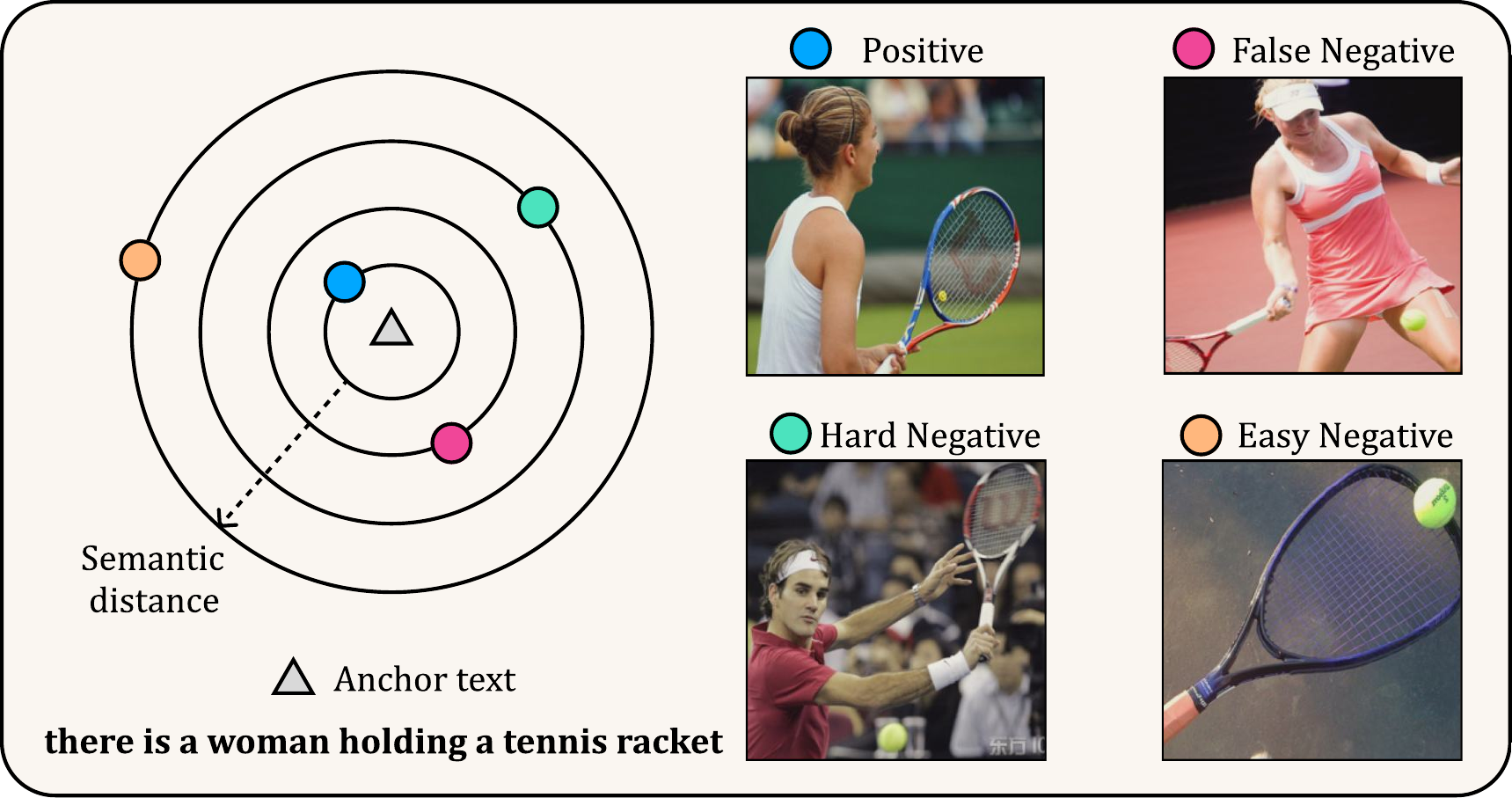}
    \caption{Illustration of semantic distance between an anchor text and multiple image samples in vision-language embedding space.} % Concentric circles denote increasing semantic distance from the anchor text embedding.}
    \label{fig:relationship}
\vspace{-0.5cm}
\end{figure}

Although effective, these hard negative sampling strategies exhibit a critical limitation: as the selection criterion favors negatives with higher semantic similarity to the anchor, the likelihood of mistakenly identifying true positives as negatives (i.e., \textit{false negatives}) correspondingly increases \citep{byun2024mafa} (see Figure~\ref{fig:relationship}). This challenge is particularly acute in VLP, where large-scale web-crawled datasets often exhibit many-to-many correspondences between images and text \citep{byun2024mafa, jiang2023vision}. Incorporating such false negatives into contrastive learning can substantially degrade representation quality by compelling the model to separate embeddings that should ideally remain close within the shared embedding space.

\begin{figure*}[t!]
    \centering
    \includegraphics[width=0.9\linewidth]{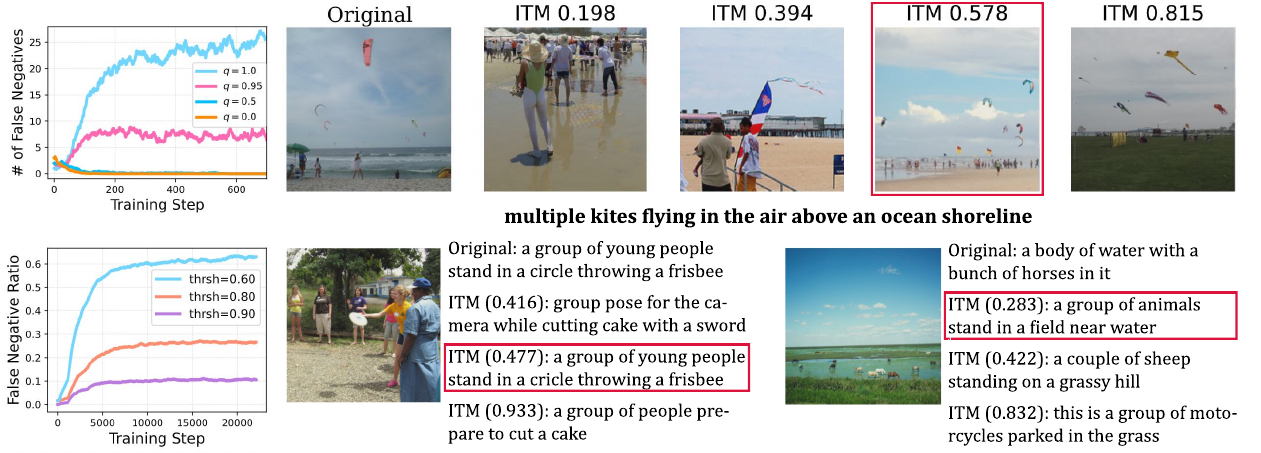} 
    \caption{\textbf{(Top Left)} Risk of false negatives for different similarity quantiles $q$ used in mini-batch construction during VLP \citep{byun2022grit}. False negatives are identified using the pretrained BLIP-129M model \citep{li2022blip}. \textbf{(Bottom Left)} False negative ratio during VLP when false negative filtering is applied using the pretrained model’s ITM score at varying thresholds. On the right, Text-to-Image \textbf{(Top Right)} and Image-to-Text \textbf{(Bottom Right)} examples are shown where the pretrained model fails to assign high ITM scores to false negatives (highlighted in red).%Instances highlighted in red represent false negatives.
    }
    \label{false_negatives_motivation}
    \vspace{-0.3cm}
\end{figure*}

To address this challenge, a few recent works attempted to mitigate this issue by leveraging strong pretrained models, either by refining the loss function to reduce the impact of false negative \citep{jiang2023vision, radenovic2023filtering} or relabeling false negatives as positives \citep{byun2024mafa, bulat2024fff}. However, these methods heavily depend on the reliability of the pretrained models and the heuristics used to identify false negatives, potentially limiting their generalizability across diverse datasets and training settings (see Appendix~\ref{appd:datacomp}). 

In this paper, we present FALCON (False-negative Aware Learning of COntrastive Negatives), a learning-based mini-batch construction strategy for VLP that adaptively balances the trade-off between informative (hard) negatives and misleading (false) negatives. By explicitly optimizing this trade-off during mini-batch construction, FALCON effectively accounts for the inevitable presence of false negatives, improving the quality of learned cross-modal representations. Our empirical results demonstrate that FALCON significantly outperforms heuristic baselines across a wide range of downstream tasks and evaluation settings. 
To summarize, our three main contributions are as follows: (i) To the best of our knowledge, this is the first \textbf{\textit{learning-based}} approach to explicitly schedule the trade-off between hard and false negatives in contrastive learning, and to empirically demonstrate its significance in vision-language pretraining. (ii) We propose a novel architecture of negative mining scheduler and an associated training scheme, both carefully designed to balance the trade-off effectively without incurring significant training overhead. (iii) We conduct experiments on three distinct vision-language frameworks (ALBEF, BLIP-2, SigLIP-2) across diverse downstream tasks and experimental settings to demonstrate the effectiveness of FALCON. Due to space constraints, the \textbf{detailed discussion on related works} is provided in the Appendix~\ref{related works}.

%% file: sec/2_motivation.tex
\section{Motivation}
In this section, we motivate the need for a scheduling strategy that dynamically balances the trade-off between hard and false negatives throughout the VLP process.

\paragraph{Optimal similarity is anchor-specific and evolves over training} To retrieve hard negatives, prior works have commonly adopted cosine similarity of embeddings as a selection criterion \citep{li2021align, li2022blip, li2023blip, byun2022grit, byun2024mafa}, selecting images or texts with high similarity to the anchor. It is widely acknowledged that there exists an optimal similarity range that facilitates effective hard negative mining while minimizing the risk of false negative inclusion \citep{wu2020conditional, robinson2020contrastive}. However, this optimal range is inherently anchor-specific and evolves throughout the training process, making it impractical to capture with a single fixed similarity level. In particular, semantically complex anchors are more difficult to align and are therefore learned more slowly, resulting in less mature and noisier embeddings even in later stages of training. For such anchors, or during the early phases of training, selection of negatives with sufficiently low similarity is necessary to reduce the risk of false negatives. In contrast, semantically simple anchors are aligned more rapidly and yield well-formed embeddings earlier in training. In these cases, or in the later stage of training, the similarity distribution of positive pairs becomes tighter and concentrated at higher values, permitting the safe mining of more similar negatives.

Consequently, VLP methods that rely on embedding similarity are exposed to a dynamic, anchor-dependent risk of false negative selection that evolves throughout the training process. 
As shown in Figure~\ref{false_negatives_motivation} (top left), the false negative rate increases substantially when highly similar samples ($q = 1.0$) are selected for mini-batch construction during vision-language pretraining. Here, $q \in [0, 1]$ denotes the quantile level used to define the hardness of negative samples within the candidate pool. Over the course of training, false negatives associated with simple semantic anchors progressively cluster in the upper quantiles of the similarity spectrum, exacerbating the risk associated with high-$q$ negative sampling (i.e., aggressive hard negative sampling). Conversely, selecting less similar samples ($q \leq 0.5$) results in a progressively lower incidence of false negatives throughout training. These findings emphasize the need for an adaptive strategy that dynamically adjusts the similarity threshold based on the anchor's semantic characteristics and the evolving state of the embedding space.

\paragraph{Adoption of pretrained models is not a complete solution} 
To address this challenge, prior works have leveraged strong pretrained models (e.g., BLIP with 129M parameters) to detect potential false negatives in image-text pairs \citep{jiang2023vision, radenovic2023filtering, byun2024mafa, bulat2024fff}. For example, MAFA~\citep{byun2024mafa} uses the Image-Text Matching (ITM) score predicted by a pretrained model to identify false negatives, relabeling them as positives when the score exceeds a predefined threshold. While this approach mitigates the nonstationarity associated with learned similarity metrics—since the pretrained model remains fixed throughout training—its effectiveness is still limited by the anchor-specific nature of the scores.
For simple descriptions or images, the set of valid positive samples is typically large and diverse, enabling pretrained models to easily assign high ITM scores to positives (and false negatives). In contrast, pretrained models struggle to generalize in more complex scenarios due to limited prior exposure, often assigning low ITM scores even to semantically aligned pairs (see Figure~\ref{false_negatives_motivation}, right).
Consequently, applying a fixed ITM threshold would be either overly conservative for simple pairs or insufficient to eliminate false negatives in complex pairs.

Furthermore, the conventional two-stage framework, which selects the most similar negatives followed by filtering with a pretrained model \citep{byun2024mafa}, is highly sensitive to hyperparameter choices, since the initial stage often includes a substantial proportion of false negatives. If the filtering threshold of the pretrained model is misspecified, training can be severely hindered by an excessively high false negative rate, reaching up to $60\%$ (see Figure~\ref{false_negatives_motivation}, bottom left).

\begin{figure*}[t!]
    \centering
    \includegraphics[width=0.8\linewidth]{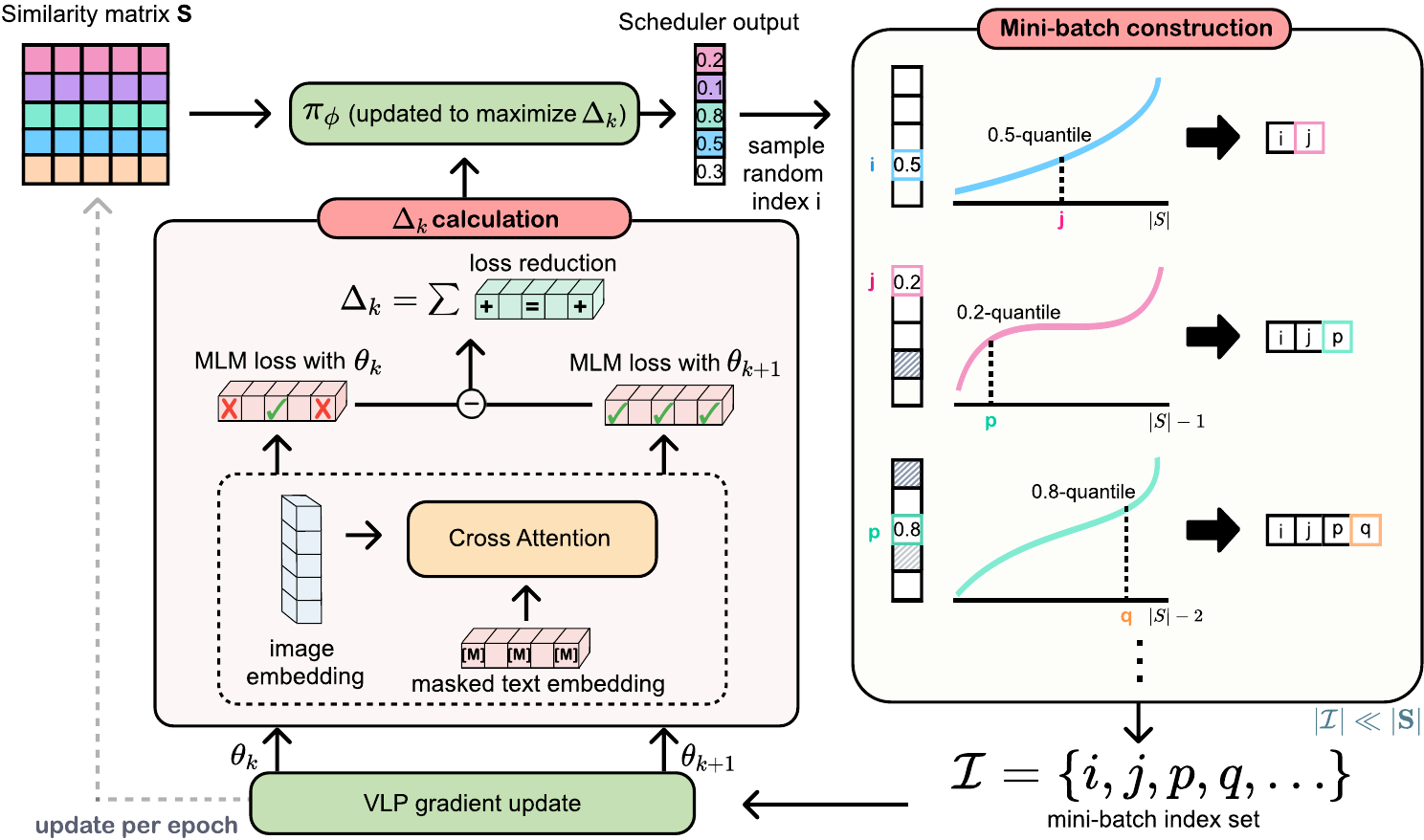}
    \caption{Overview of FALCON, a learning-based mini-batch construction strategy for VLP. Starting from a randomly selected anchor, the scheduler $\pi_\phi$ predicts hardness quantile values to iteratively sample additional candidates, forming a mini-batch index set $\mathcal{I}$. This batch is used to update the VLP model. The reduction in masked language modeling loss $\mathcal{L}_\text{MLM}$ serves as a proxy for enhanced cross-modal alignment, providing a learning signal that guides $\pi_\phi$ toward constructing more informative mini-batches in subsequent training steps.
    }
    \label{main_architecture}
    \vspace{-0.25cm}
\end{figure*}

%% file: sec/3_main.tex
\section{Main Method}
Recent VLP models \citep{li2021align, li2022blip, li2023blip} are typically optimized using a combination of contrastive and generative objectives, as expressed by the following loss formulation:
\begin{align}
\begin{split}
\min_\theta \mathcal{L}_\text{VLP}(\theta):= &~ \mathbb{E}_{(V,T)\sim \mathcal{V}, \mathcal{T}}\big[\mathcal{L}_{\text{ITC}}(V,T;\theta)\\
    +&\mathcal{L}_{\text{ITM}}(V,T;\theta)+\mathcal{L}_{\text{MLM}}(V,T;\theta)\big].
    \label{eq:VLPLoss}
\end{split}
\end{align}
where $\theta$ denotes the parameters of the VLP model, $\mathcal{L}_{\text{ITC}}$, $\mathcal{L}_{\text{ITM}}$, and $\mathcal{L}_{\text{MLM}}$ correspond to the image-text contrastive (ITC) loss, image-text matching (ITM) loss, and masked language modeling (MLM) loss, respectively. In particular, ITC serves as a contrastive loss that aligns embeddings for matched image-text pairs (positive pair) while pushing apart mismatched pairs (negative pair) in the embedding space. Complementarily, ITM is a binary matching objective that distinguishes positive and negative pairs, enhancing cross-modal alignment. The MLM loss can be replaced with alternative generative objectives, e.g., the BLIP family~\citep{li2022blip,li2023blip}.

Due to the inclusion of contrastive objectives (ITC, ITM), the quality of the learned embedding space is strongly influenced by the hardness of the negative samples \citep{schroff2015facenet, li2021align, li2022blip}. Traditionally, each instance within a mini-batch serves as an anchor, with all remaining instances in the same batch acting as its negative samples. As such, the composition of the mini-batch critically influences both the quality and the difficulty of the negative samples \citep{he2020momentum, li2021align, byun2022grit}. However, as discussed in the previous section, existing strategies for mini-batch construction have not sufficiently accounted for the anchor-dependent and training-dependent variability in optimal negative sample hardness.

In this section, we present FALCON, a learning-based mini-batch construction strategy that adaptively balances hard and false negatives by optimizing a scheduler $\pi_\phi$ to determine an appropriate negative hardness level for each anchor during mini-batch construction in VLP.

\subsection{Construction of Mini-batch using Negative Mining Scheduler}
While the VLP loss~\eqref{eq:VLPLoss} was initially proposed with conventional uniform sampling of a mini-batch $(V, T)$, GRIT-VLP~\citep{byun2022grit} modifies the sampling procedure by introducing a mini-batch grouping strategy that iteratively selects the most semantically similar sample to the most recently chosen sample, thereby promoting effective hard negative mining within each batch. For scalability, the dataset is partitioned into multiple localized search spaces $\{M\}$ and the most similar sample is searched within the search space. However, GRIT-VLP does not account for the heightened risk of false negatives associated with mining increasingly harder negatives, which substantially limits the effectiveness of hard negative sampling; for example, excessively large search space size $|M|$ can lead to performance degradation due to the elevated likelihood of false negatives (see Appendix~\ref{appd:baseline sweep}).

To this end, we extend the strategy of GRIT-VLP by selecting the next sample based on a specified level of similarity to the most recently chosen sample, where the similarity level is determined by a scheduler $\pi_\phi$ rather than fixed to the most similar sample. By jointly optimizing the scheduler during training, we aim to adaptively balance the trade-off between hard and false negatives by dynamically selecting the appropriate similarity level on a per-instance basis during mini-batch construction.

Specifically, the scheduler $\pi_\phi$ predicts a hardness quantile value $q \in [0, 1]$ for each anchor. When the scheduler determines that an anchor would benefit from more informative and challenging negatives under the current state of the VLP model, it samples a quantile value closer to 1, thereby selecting harder negatives. Note that a scheduler consistently producing $q=1$ effectively reduces to the GRIT-VLP strategy. Conversely, when the potential risk of sampling false negatives is high, the scheduler assigns a lower quantile value, favoring negatives that are semantically dissimilar and thus less likely to overlap with true positives. 

A detailed mini-batch construction process is illustrated in Figure~\ref{main_architecture}, with corresponding pseudo-code provided in Appendix~\ref{algorithms}. The process begins by uniformly selecting an initial sample from the local search space. Then, a new sample is chosen according to the hardness quantile value $q$ predicted by the scheduler $\pi_\phi$, and added to the mini-batch. This procedure is applied recursively until a mini-batch of size $B$ is formed: at each step, a new sample is selected using the hardness quantile, excluding previously selected samples from further selection. Once the mini-batch is constructed, it is used to update the VLP model parameterized by $\theta$. As the size of search space $M$ is larger than the mini-batch size $B$, the procedure of constructing mini-batches and updating the VLP model is repeated iteratively until all candidates within the search space have been utilized. Through this tactical grouping strategy, negative samples that align with the desired hardness level for each anchor are progressively incorporated into the VLP process (more details in Appendix~\ref{appd:details_minibatch}).

\begin{figure*}[t!]
    \centering      
    \includegraphics[width=1.0\linewidth]{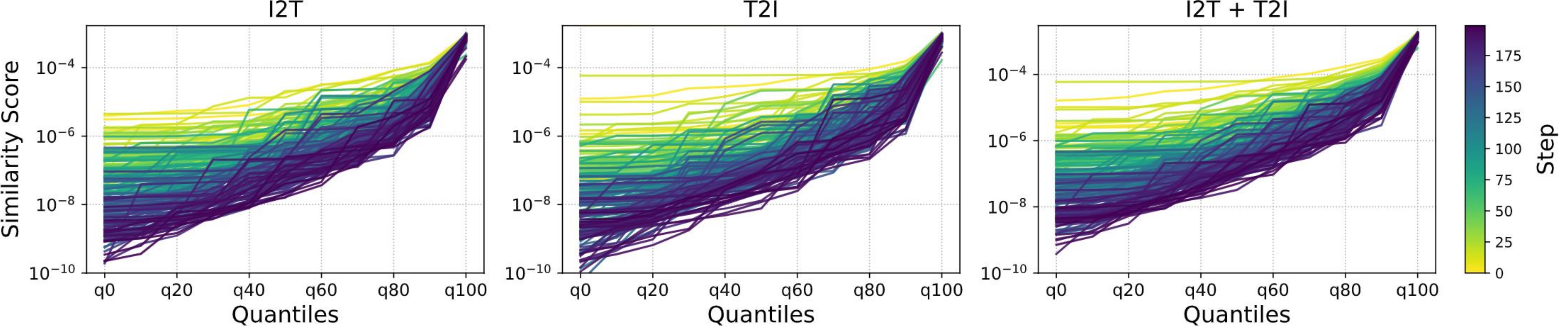}
    \caption{Visualization of normalized cosine similarity distributions during training. Each plot shows the similarity at different quantile levels for I2T (left), T2I (center), and I2T+T2I (right), with the color bar indicating training progression. Step values are shown in thousands (K).}
    \label{cosine_similarity_throughout_training}
    \vspace{-0.5cm}
\end{figure*}
\subsection{Design of Negative Mining Scheduler}
To enable the scheduler $\pi_\phi$ to adaptively select negative samples with appropriate hardness for each anchor instance during mini-batch construction, we provide it with a matrix of similarity distributions computed over samples within the search space. Instead of feeding the anchor instances directly into the scheduler, we use the distribution of similarities between the anchor and the candidate samples, represented by the corresponding rows of the similarity distribution matrix. This serves as an efficient surrogate for estimating the similarity levels required to identify suitable hard negatives.

Specifically, we construct a unified similarity matrix $\mathbf{S}$ by adding the image-to-text (I2T) and text-to-image (T2I) pairwise similarity matrices among candidates within the local search space. These matrices are computed from the \texttt{[CLS]} embeddings of image and text representations and contain rich information about the semantic alignment between modalities in the representation space. Following prior works, we leverage cached \texttt{[CLS]} embeddings maintained in a queue to compute $\mathbf{S}$, thereby avoiding additional forward passes~\citep{he2020momentum, li2021align, byun2022grit}.

By integrating both I2T and T2I similarities into a single matrix, the scheduler $\pi_\phi$ can operate in a direction-agnostic manner, thereby eliminating the need to alternate between modalities during negative sample selection. Figure~\ref{cosine_similarity_throughout_training} visualizes similarity distributions of both I2T and T2I; as the two matrices exhibit comparable scales, we did not employ a more complex combination strategy. 

To facilitate efficient learning of similarity distributions, we select $m$ evenly spaced quantiles from each row instead of using the full set of raw similarity values, yielding a compact matrix of size $|M| \times m$ where $m \ll |M|$. 
Afterwards, row-wise normalization using softmax is applied to minimize the impact of changing scale of pairwise similarities over training (see Figure~\ref{cosine_similarity_throughout_training}). We call this normalized similarity distribution matrix $\widehat{\mathbf{S}}$.

\begin{figure*}[t!]
    \centering      
    \includegraphics[width=0.9\linewidth]{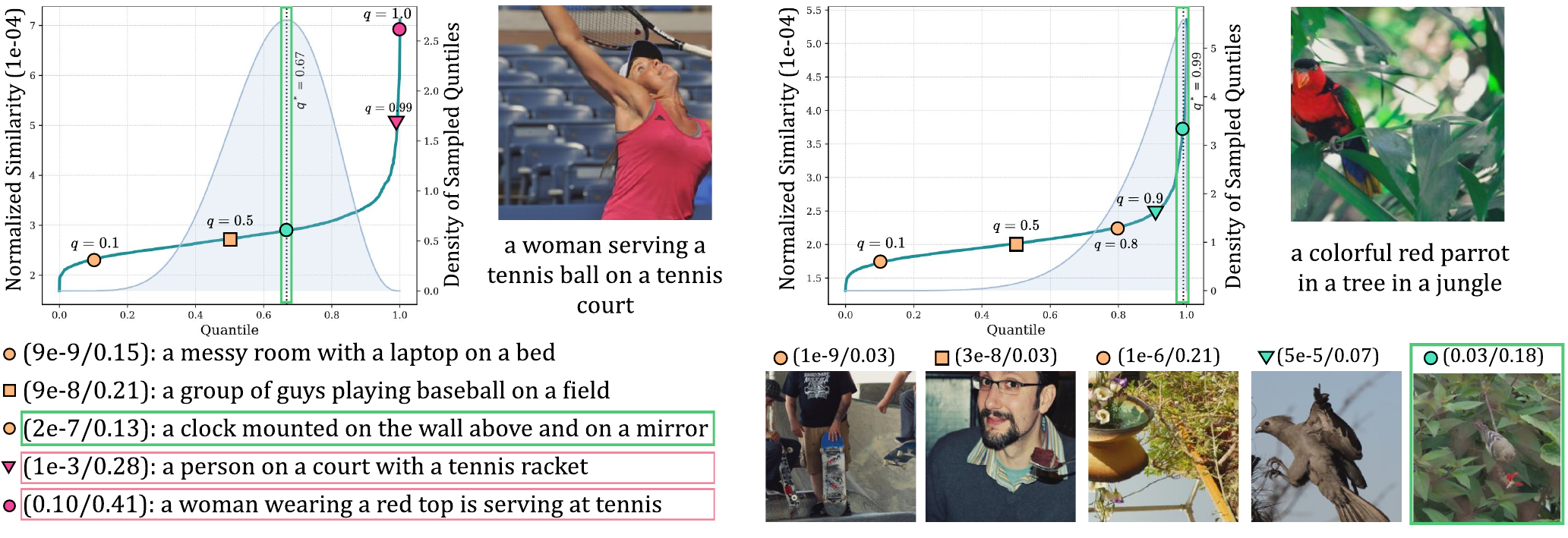}
    \caption{Image-to-Text (Left) and Text-to-Image (Right) examples illustrating FALCON’s quantile-based negative sampling strategy. For each anchor query (shown adjacent to each plot), the normalized similarity distribution $\widehat{\mathbf{S}}$ over the candidate set is displayed alongside the scheduler’s predicted distribution on quantiles (blue-shaded density curve). Sampled negatives are annotated with their (one-way similarity / pretrained ITM score), and color-coded by hardness level as defined in Figure~\ref{fig:relationship}.}
    \label{falcon_negative_sample_illust}
    \vspace{-0.3cm}
\end{figure*}

The scheduler $\pi_\phi$ is implemented as a lightweight 4-layer residual MLP that maps $\widehat{\mathbf{S}}$ to the parameters $(\alpha, \beta)$ of Beta distributions, which model the desired hardness (similarity quantile) of negative samples for each anchor. To ensure permutation equivariance over rows without relying on heavier architectures~\citep{zaheer2017deep,lee2019set,jaegle2021perceiver}, we sort the rows of $\widehat{\mathbf{S}}$ prior to inputting it into the MLP~\citep{kimura2024permutation}. This design choice offers great computational and memory efficiency while maintaining sufficient expressivity for the scheduling task. Figure~\ref{falcon_negative_sample_illust} presents qualitative examples of FALCON’s quantile-based sampling. 

\subsection{Training Negative Mining Scheduler}
The goal of VLP is to learn a unified representation space that effectively aligns visual and textual modalities, thereby enabling generalization across a wide range of downstream multimodal tasks. To support this objective, we design the training signal for $\pi_\phi$ to encourage mini-batch constructions that enhance cross-modal alignment within the learned representation space. 

Specifically, we formulate the objective as \textbf{maximizing the reduction in the masked language modeling loss} $\mathcal{L}_\text{MLM}$ measured before and after updating the VLP model with a mini-batch constructed by $\pi_\phi$, while keeping the masking identical for loss evaluations.\footnote{For models in the BLIP family, the improvement can alternatively be measured using their generative objectives, such as the language modeling (LM) loss~\citep{li2022blip} or the image-text generation (ITG) loss~\citep{li2023blip}.} This change in $\mathcal{L}_\text{MLM}$ serves a proxy for improvements in the model’s ability to integrate visual context into language understanding via cross-attention, thus reflecting more refined semantic alignment between the two modalities. In contrast, training $\pi_\phi$ to minimize other objectives (e.g., $\mathcal{L}_\text{ITC}$, $\mathcal{L}_\text{ITM}$) that rely on negative samples undermines VLP training by encouraging the exploitation of trivial negatives, which is empirically validated in our experiments. 

Formally, we aim to update our scheduler $\pi_{\phi_k}$ at the $k$-th gradient update step as:
\begin{gather}
    \phi_k = \argmax_{\phi_k} \mathbb{E}_{(V,T)\sim\pi_{\phi_k}}[\Delta_k^{V,T}],\\
    \text{where}\quad \Delta_k^{V,T}:=\mathcal{L}_{\text{MLM}}(V,T;\theta_k)-\mathcal{L}_{\text{MLM}}(V,T;\theta_{k+1}) \nonumber
\end{gather}
where $(V,T)\sim\pi_{\phi_k}(\cdot\,\vert\, \widehat{\mathbf{S}})$ denotes a mini-batch constructed using the current scheduler $\pi_\phi$ given similarity matrix $\mathbf{S}$. To optimize this objective, we use log-derivative trick~\citep{sutton1999policy} and update as:
\begin{equation}
    \phi_{k+1} = \phi_k +\gamma\cdot \mathbb{E}_{\pi_{\phi_k}}[\Delta_k^{V,T}\cdot\nabla_{\phi_k}\log\pi_{\phi_k}(V,T\,\vert\,\widehat{\mathbf{S}})].
\end{equation}
where $\gamma$ denotes the step size. By leveraging the progressively optimized scheduler $\pi_{\phi_k}$, the VLP model is guided to learn a unified semantic representation space that robustly aligns visual and textual modalities, incorporating appropriately challenging negative samples. We summarize the training loop of FALCON in Appendix~\ref{algorithms}. 

%% file: sec/4_exp.tex
\section{Experiments}
\label{experiments}
We begin by evaluating the effectiveness of FALCON in comparison to heuristic negative mining approaches in Section~\ref{exp:comparison_to_heruistics}. Then we assess the compatibility of FALCON with BLIP-2 and SigLIP-2 in \Cref{exp:compatibility_with_BLIP,exp:compatibility_with_SIGLIP}, respectively. Finally, we present additional analyses and ablation studies to further support our method in Section~\ref{exp:ablation_FALCON}. Experimental details are provided in Appendix~\ref{experimental details} and additional studies are provided in Appendix~\ref{appd:addtional_exp}.

\subsection{Comparison to Heuristic Negative Mining Methods}
\label{exp:comparison_to_heruistics}

\paragraph{Comparison across downstream tasks} To compare FALCON with existing heuristic negative mining methods, we pretrained all models on the MSCOCO dataset \citep{lin2014microsoft} and evaluated them on three downstream tasks (IRTR \citep{frome2013devise}, VQA \citep{antol2015vqa}, NLVR \citep{suhr2018corpus}). As MSCOCO is a human-curated dataset with minimal annotation noise compared to web-crawled datasets \citep{sharma2018conceptual, ordonez2011im2text}, it allows for an isolated assessment of the effectiveness of negative sampling strategies, minimizing confounding effects from noisy image-text alignments. In addition to standard baselines in prior work, we introduce four heuristic scheduling strategies for comparative analysis. $q=0.0$ and $q=0.5$ represent fixed-hardness negative sampling heuristics, where negatives are selected at constant hardness levels. Note GRIT-VLP corresponds to the $q=1.0$ setting. Beyond these, we define two dynamic scheduling baselines: Progressive-Hardening, which gradually increases the sampled hardness level over training epochs, and Progressive-Softening, which decreases it. These schedules simulate curriculum-like strategies for hardness adjustment. 
\begin{table*}[t!]
\centering
\caption{Performance Comparison of FALCON with existing heuristic negative mining methods on IRTR, VQA and NLVR2. All models are pretrained on the MSCOCO dataset~\citep{lin2014microsoft}. \textbf{Bold} denotes the best result.}
% \resizebox{\linewidth}{!}{
\begin{tabular}{l | ccc ccc | cc | cc}
\toprule
\multirow{2}{*}{\textbf{Method}} 
& \multicolumn{3}{c}{\textbf{Text Retrieval}} 
& \multicolumn{3}{c|}{\textbf{Image Retrieval}} 
& \multicolumn{2}{c|}{\textbf{NLVR2}}
& \multicolumn{2}{c}{\textbf{VQA}} \\
& R@1 & R@5 & R@10 
& R@1 & R@5 & R@10 
& dev & test-P 
& test-dev & test-std \\
\midrule
ALBEF~\citep{li2021align}       & 55.60 & 81.92 & 90.10 & 41.16 & 70.63 & 80.81 & 72.98 & 73.61 & 70.46 & 70.72 \\
GRIT-VLP~\citep{byun2022grit}   & 60.60 & 83.52 & 89.14 & 44.61 & 69.54 & 77.67 & 74.63 & 75.26 & 71.04 & 71.22 \\
DiHT~\citep{radenovic2023filtering} & 54.72 & 78.64 & 84.00 & 40.53 & 65.20 & 74.22 &  73.08 & 74.12 & 70.74 & 71.07 \\
SRCL~\citep{jiang2023vision}    & 54.52 & 79.24 & 85.42 & 41.25 & 66.36 & 75.04 & 73.27 & 74.28 & 70.77 & 71.01 \\
MAFA~\citep{byun2024mafa}       & 60.96 & 83.24 & 89.62 & 44.77 & 69.49 & 77.96 & 75.16 & 75.13 & 71.13 & 71.22 \\
$q=0.0$    & 18.92 & 55.30 & 72.90 & 20.18 & 58.07 & 74.73 & 72.32 & 73.42 & 69.97 & 70.30 \\
$q=0.5$    & 58.80 & 84.16 & 91.28 & 44.73 & 73.43 & 83.26 & 73.98 & 74.41 & 71.00 & 71.31 \\
Progressive-Hardening    & 58.30 & 84.54 & 90.94 & 44.21 & 73.39 & 82.97 & 73.42 & 74.24 & 70.95 & 71.15 \\
Progressive-Softening    & 59.08 & 84.76 & 91.74 & 44.24 & 73.59 & 82.74 & 73.58 & 74.80 & 70.93 & 71.17 \\
\midrule
FALCON                            & \textbf{62.28} & \textbf{86.18} & \textbf{92.30} & \textbf{46.18} & \textbf{74.65} & \textbf{83.58} &  \textbf{75.17} & \textbf{75.47} & \textbf{71.24} & \textbf{71.36} \\
\bottomrule
\end{tabular}
% }
\label{table:supple-overall-baseline}
\end{table*}

Table~\ref{table:supple-overall-baseline} demonstrates that FALCON achieves significantly better performance compared to heuristic negative mining methods. Notably, hard negative sampling approaches that do not explicitly address false negatives (GRIT-VLP, DiHT) suffer from degraded performance, as false negatives introduce conflicting supervision signals. Meanwhile, methods that attempt to mitigate false negatives by leveraging pretrained models (MAFA, SRCL) also face inherent limitations. We believe that these approaches, which rely on fixed similarity thresholds or static filters derived from pretrained representations, are insufficient to capture the dynamic nature of optimal similarity throughout training (see Appendix~\ref{appd:datacomp}). In contrast, Figure~\ref{hard_negative_quantile_scheduling} shows that FALCON adaptively adjusts its strategy to reflect the temporally evolving optimal similarity range, enabling more effective hard negative selection. Early in training, FALCON predominantly samples high-quantile negatives to construct challenging mini-batches that accelerate representation learning. As the embedding space gradually matures, the likelihood of encountering false negatives within the search space increases. Consequently, FALCON adaptively lowers its target quantile to mitigate the adverse impact of these false negatives. This adaptive behavior enables FALCON to consistently outperform baseline methods across all stages of training (visualized in Appendix~\ref{appd:compare_epoch}).
\vspace{-0.3cm}

\begin{figure}[t!]
    \centering
    \includegraphics[width=0.85\columnwidth]{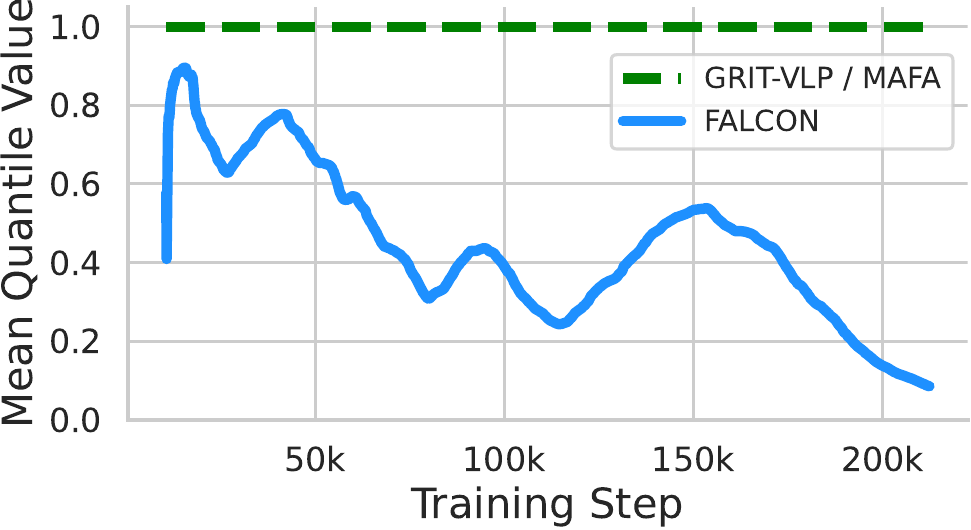}
    \caption{Visualization of hard negative scheduling of FALCON and baselines. MAFA additionally relies on a pretrained model to relabel false negatives as positives.}
    \label{hard_negative_quantile_scheduling}
    \vspace{-0.5cm}
\end{figure}

\begin{figure*}[t!]
    \centering
    % Subfigure: Table
    \begin{subfigure}[]{0.5\textwidth}
        \centering
        \includegraphics[width=\linewidth]{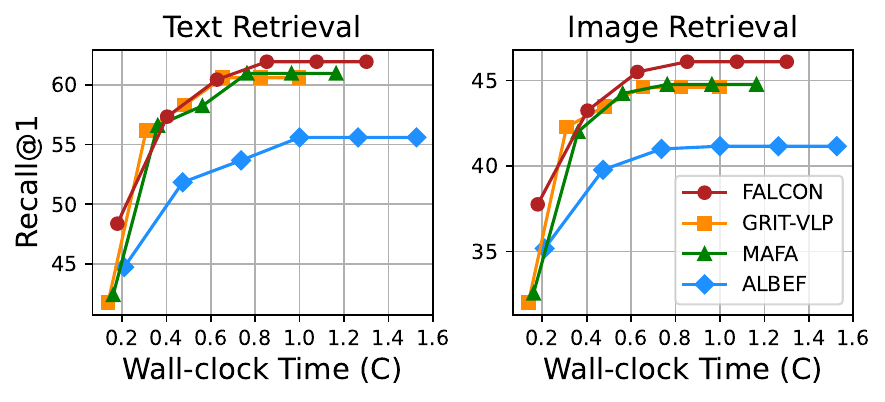}
    \end{subfigure}
    \hspace{1cm}
    \begin{subfigure}[]{0.38\textwidth}
        \centering
        % \resizebox{1.0\textwidth}{!}{
        \begin{tabular}{c c}
		\toprule
		\textbf{Method} & Time for convergence \\
		\midrule
		ALBEF & 1.0C \\
		GRIT-VLP & 0.65C \\
		MAFA & 0.76C \\
        \midrule
        FALCON  & \multirow{1}{*}{0.83C}\\
		\bottomrule
	\end{tabular}%}
    \end{subfigure}
    \caption{Comparison of FALCON and baselines against relative wall clock time for Recall@1 on the IRTR task. 1C denotes the wall clock time required for ALBEF to reach convergence, defined as the point where performance improvement stops.}
    \label{fig:combined_pg}
    \vspace{-0.3cm}
\end{figure*}

\paragraph{Comparison on relative wall clock time} Figure~\ref{fig:combined_pg} presents the performance of FALCON and baselines against relative wall clock time,  where 1C denotes the total time required for ALBEF to converge. ALBEF incurs the highest per-epoch training time due to its reliance on a momentum model for generating soft labels, whereas the other methods utilize computationally efficient soft pseudo targets. Although FALCON is architecturally optimized to minimize computational overhead, it requires additional forward passes and updates for the scheduler network, leading to a moderately higher per-epoch cost compared to MAFA and GRIT-VLP. Nevertheless, FALCON achieves better performance relative to wall clock time, highlighting the effectiveness and importance of dynamically balancing the trade-off between hard and false negatives during vision-language pretraining.

\vspace{-0.2cm}
\paragraph{Comparison on Standard 4M Datasets} 
To further assess the robustness of FALCON in mitigating false negatives, we conducted experiments on large-scale web-crawled image-text datasets, including Conceptual Captions \citep{sharma2018conceptual} and SBU Captions \citep{ordonez2011im2text}, in addition to Visual Genome \citep{krishna2017visual} and MSCOCO, following the protocol introduced in ALBEF \citep{li2021align}. This evaluation assesses whether FALCON can effectively mitigate false negatives on large noisy web-crawled dataset. Table~\ref{table:4M-noisy} shows that FALCON achieves the best overall performance. However, the performance gap is less pronounced compared to the results in Table~\ref{table:supple-overall-baseline}. We attribute this to significant noise and semantic misalignment in the original captions, which can hinder accurate estimation of the tradeoff between hard and false negatives, thereby making the learning process more complex. Additional experiments on the DataComp benchmark (Appendix~\ref{appd:datacomp}) and a BLIP-captioned version of the dataset (Appendix~\ref{appd:blip_captioned}) reveal an increased performance gap, supporting these observations.

% demonstrating that false-negative aware mini-batch construction yields strong vision–language alignment even with large noisy web-crawled dataset.
% The pretrained model is directly evaluated on the MSCOCO for zero-shot retrieval, and we fine-tune the pretrained model with its training dataset for the fine-tuned setting. As shown in Table~\ref{table:4M-noisy}, 

\begin{table*}[t!]
\centering
\caption{Zero-shot and fine-tuned image-text retrieval performance comparison on MSCOCO. \textbf{Bold} denotes the best result among models pretrained with 4M dataset. See Table~\ref{tab:vltasks} in Appendix for more results.}
\setlength\tabcolsep{4.5pt}
\resizebox{\linewidth}{!}{
\begin{tabular}{l | c | ccc ccc | ccc ccc}
\toprule
\multicolumn{1}{c|}{\multirow{3}{*}{\textbf{Method}}} & \multirow{3}{*}{\#Images} & \multicolumn{6}{c|}{\textbf{Zero-shot}} & \multicolumn{6}{c}{\textbf{Fine-tuned}} \\
\multicolumn{1}{c|}{} & & \multicolumn{3}{c}{Text Retrieval} & \multicolumn{3}{c|}{Image Retrieval} & \multicolumn{3}{c}{Text Retrieval} & \multicolumn{3}{c}{Image Retrieval} \\
\multicolumn{1}{c|}{} & & R@1 & R@5 & R@10 & R@1 & R@5 & \multicolumn{1}{c|}{R@10} & R@1 & R@5 & R@10 & R@1 & R@5 & R@10 \\
\midrule
ImageBERT~\cite{qi2020imagebert} & 6M   & 44.0 & 71.2 & 80.4 & 32.3 & 59.0 & 70.2 & 66.4 & 89.8 & 94.4 & 50.5 & 78.7 & 87.1 \\
UNITER~\cite{chen2020uniter} & 4M   & 64.1 & 87.7 & 93.3 & 48.8 & 76.7 & 85.8 & 65.7 & 88.6 & 93.8 & 52.9 & 79.9 & 88.0 \\
ViLT~\cite{kim2021vilt} & 4M     & 56.5 & 82.6 & 89.6 & 40.4 & 70.0 & 81.1 & 61.5 & 86.3 & 92.7 & 42.7 & 72.9 & 83.1 \\
ALBEF~\cite{li2021align} & 4M     & 68.7 & 89.5 & 94.7 & 50.1 & 76.4 & 84.5 & 73.1 & 91.4 & 96.0 & 56.8 & 81.5 & 89.2 \\
TCL~\cite{yang2022vision} & 4M     & 71.4 & 90.8 & 95.4 & 53.5 & 79.0 & 87.1 & 75.6 & 92.8 & 96.7 & 59.0 & 83.2 & 89.9 \\
ALIGN~\cite{jia2021scaling} & 1.2B & 58.6 & 83.0 & 89.7 & 45.6 & 69.8 & 78.6 & 77.0 & 93.5 & 96.9 & 59.9 & 83.3 & 89.8 \\
GRIT-VLP~\cite{byun2022grit} & 4M     & - & - & - & - & - & - & 76.6 & 93.4 & 96.9 & 59.6 & 83.3 & 89.9 \\
MAFA~\cite{byun2024mafa} & 4M     & 72.6 & 91.3 & 95.6 & 53.9 & 79.6 & \textbf{87.7} & 78.0 & 93.4 & 96.9 & 61.2 & \textbf{83.9} & \textbf{90.2} \\
\midrule
FALCON & 4M & \textbf{74.1} & \textbf{91.5} & \textbf{95.9} & \textbf{54.8} & \textbf{79.8} & 87.6 & \textbf{78.7} & \textbf{94.0} & \textbf{97.0} & \textbf{61.5} & 83.6 & \textbf{90.2} \\
\bottomrule
\end{tabular}
}
\label{table:4M-noisy}
\vspace{-0.3cm}
\end{table*}

\begin{table}[t!]
    \centering
    \caption{Performance Comparison under the BLIP-2 framework. See \Cref{table:blip-detail} in Appendix for full results.}
    \resizebox{\linewidth}{!}{
    \begin{tabular}{c | cc cc | c c c}
        \toprule
        \multirow{3}{*}{\textbf{Method}} & \multicolumn{4}{c|}{\textbf{Stage-1}} & \multicolumn{3}{c}{\textbf{Stage-2}} \\
        & \multicolumn{2}{c}{\textbf{COCO R@1}} & \multicolumn{2}{c|}{\textbf{Flickr R@1}} & \textbf{VQA2} & \textbf{OKVQA} &  \textbf{Capt.} \\
        & TR & IR & TR & IR & val & test & SPICE \\
        \midrule
        BLIP-2          & 75.22 & 57.98 & 90.10 & 77.48 & 42.46 & 17.94 &  \textbf{19.5} \\
        + GRIT-VLP      & 73.90 & 57.47 & 90.40 & 77.28 & 39.91 & 15.92 & 19.4 \\
        + MAFA          & 74.21 & 57.94 & 90.30 & 77.32 & 41.12 & 18.54 & 19.3 \\
        + FALCON        & \textbf{75.56} & \textbf{58.52} & \textbf{90.90} & \textbf{77.72} & \textbf{42.67} & \textbf{20.96} &  19.4 \\
        \bottomrule
    \end{tabular}
    }
    \label{table:blip}
    \vspace{-0.3cm}
\end{table}

\subsection{Compatibility with BLIP-2}
\label{exp:compatibility_with_BLIP}
To evaluate the applicability of FALCON to other vision-language frameworks, we examined its compatibility with BLIP-2 \citep{li2023blip}, a BLIP-family model demonstrating strong performance across various multimodal tasks. Unlike ALBEF, BLIP-2 replaces the masked language modeling (MLM) loss with image-grounded text generation (ITG) loss as its generative objective. Accordingly, we adopted the ITG loss as a proxy for cross-modal alignment when applying FALCON to BLIP-2. We pretrained all models on the MSCOCO dataset. As shown in Table~\ref{table:blip}, FALCON yields significant performance gains within the BLIP-2 architecture over most of the tasks, indicating that its benefits extend to vision-language models built on alternative generative objectives that jointly leverage text and image encoder information.

\begin{table}[t!]
\centering
\caption{Retrieval Performance of SigLIP-2 and SigLIP-2 + FALCON on MSCOCO.}
\resizebox{\linewidth}{!}{
\begin{tabular}{c 
                S[table-format=2.2] S[table-format=2.2]
                S[table-format=2.2] S[table-format=2.2]
                S[table-format=2.2] S[table-format=2.2]}
\toprule
\multirow{2}{*}{\textbf{Method}}
  & \multicolumn{3}{c}{\textbf{Text Retrieval}} 
  & \multicolumn{3}{c}{\textbf{Image Retrieval}} \\
  \cmidrule(lr){2-4} \cmidrule(lr){5-7}
  & {R@1} & {R@5} & {R@10} & {R@1} & {R@5} & {R@10} \\
\midrule
  SigLIP-2 & 69.96 & 89.72 & 94.12 & \textbf{54.21} & 78.52 & 86.07 \\  
  + FALCON & \textbf{72.96} & \textbf{91.00} & \textbf{95.44} & 54.15 & \textbf{78.78} & \textbf{86.49} \\
\bottomrule
\end{tabular}
}
\label{tab:siglip2}
\vspace{-0.2cm}
\end{table}

\subsection{Compatibility with SigLIP-2}
\label{exp:compatibility_with_SIGLIP}
Recently, SigLIP-2 \citep{tschannen2025siglip} extended the CLIP training paradigm by introducing generative objectives in vision-language alignment (image captioning, grounded captioning, and automatic referring expression prediction). These objectives are implemented by attaching a transformer decoder to the vision encoder representation. We adopted these generative losses as a proxy signal when applying FALCON to SigLIP-2. As shown in Table~\ref{tab:siglip2}, applying FALCON to SigLIP-2 yields significant gains in image-to-text retrieval, but marginal gains in text-to-image retrieval. This asymmetry arises because the auxiliary losses are computed solely through the vision encoder and transformer decoder, whereas the text encoder does not participate in these generative tasks. As a result, FALCON’s scheduling is primarily guided by improvements on the vision side. These findings suggest that, to effectively apply FALCON, the proxy for cross-modal alignment should integrate signals from both the vision and text encoders, which remains a limitation of our work.

\subsection{Ablation Studies}
\label{exp:ablation_FALCON}
Table~\ref{tab:ours_ablation} (top) illustrates the impact of search space size on FALCON. It shows that increasing the search space size initially leads to performance improvements, as it enables the scheduler to more accurately select negative samples at the predicted target hardness quantile. However, beyond a certain search space size, the performances are comparable to each other. This finding suggests that FALCON is robust to large search space sizes and varying distributions of hard and false negatives, as it effectively manages the trade-off between them. In contrast, baseline methods show performance degradation with larger search spaces, primarily due to an increased risk of false negatives (see Appendix~\ref{appd:baseline sweep}).

Table~\ref{tab:ours_ablation} (middle) demonstrates the impact of training objectives of $\pi_\phi$ as proxies for cross-modal alignment improvement. The results indicate that incorporating contrastive objectives ($\mathcal{L}_\text{ITC}$ and $\mathcal{L}_\text{ITM}$) degrades VLP performance, likely due to their tendency to exploit trivial (i.e., easy) negatives. In contrast, $\mathcal{L}_\text{MLM}$ proves to be a more effective proxy, as it remains robust to the hardness of negatives in the mini-batch.

\begin{table}[t!]
\centering
\caption{Ablation study analyzing the impact of search space size, training objectives, and instance-level scheduling of FALCON.}
\resizebox{\linewidth}{!}{
\begin{tabular}{c 
                S[table-format=2.2] S[table-format=2.2]
                S[table-format=2.2] S[table-format=2.2]
                S[table-format=2.2] S[table-format=2.2]}
\toprule
\multirow{2}{*}{\textbf{Setting}}
  & \multicolumn{3}{c}{\textbf{Text Retrieval}} 
  & \multicolumn{3}{c}{\textbf{Image Retrieval}} \\
  \cmidrule(lr){2-4} \cmidrule(lr){5-7}
  & {R@1} & {R@5} & {R@10} & {R@1} & {R@5} & {R@10} \\
\midrule
  $\vert M \vert=480$     & 58.48 & 84.70 & 91.54 & 44.75 & 73.79 & 82.99 \\
  $\vert M \vert=5664$    & 61.72 & \textbf{86.28} & \textbf{92.78} & \textbf{46.19} & 74.56 & 83.94 \\
  $\vert M \vert=28320$   & \textbf{61.94} & 85.94 & 92.58 & 46.10 & \textbf{74.61} & \textbf{83.96} \\
\midrule
  $\mathcal{L}_\text{ITC}+\mathcal{L}_\text{ITM}$ & 57.64 & 84.24 & 91.32 & 43.62 & 73.11 & 82.86 \\
  $\mathcal{L}_\text{ITC}+\mathcal{L}_\text{ITM}+\mathcal{L}_\text{MLM}$                        & 57.80 & 84.36 & 91.58 & 44.29 & 73.15 & 82.88 \\
  $\mathcal{L}_\text{MLM}$ 
& \textbf{61.72} & \textbf{86.28} & \textbf{92.78} & \textbf{46.19} & \textbf{74.56} & \textbf{83.94} \\
\midrule
  Batch-level & 58.78 & 84.08 & 90.72 & 44.47 & 73.10 & 82.52 \\  
  Instance-level & \textbf{61.72} & \textbf{86.28} & \textbf{92.78} & \textbf{46.19} & \textbf{74.56} & \textbf{83.94} \\
\bottomrule
\end{tabular}
}
\label{tab:ours_ablation}
\vspace{-0.3cm}
\end{table}

Table~\ref{tab:ours_ablation} (bottom) evaluates the impact of enabling the scheduler to dynamically assign similarity level for each anchor instance. In the \textit{Batch-level} setting, the scheduler $\pi_\phi$ selects a single similarity level shared across all anchors to construct a mini-batch used for vision-language pretraining. In contrast, the \textit{Instance-level} setting assigns a distinct similarity level to each anchor instance to form the mini-batch. The resulting performance gap highlights the importance of instance-specific similarity selection, demonstrating that the optimal similarity level is anchor-dependent and should be adaptively determined during vision-language pretraining.

%% file: sec/5_conclusion.tex
\section{Conclusion}
\label{conclusion}
In this paper, we addressed a fundamental challenge in VLP: balancing the trade-off between informative hard negatives and misleading false negatives. FALCON is a learning-based mini-batch construction strategy that dynamically schedules negative sampling to optimize this trade-off. Experimental results demonstrate that FALCON significantly outperforms heuristic negative mining strategies across various experimental settings. A discussion of limitations and future directions is provided in Appendix~\ref{appd:limitations}.

\section{Acknowledgments}
This work was partly supported by Institute of Information $\&$ Communications Technology Planning $\&$ Evaluation (IITP) grant funded by the Korea government (MSIT) (No. RS-2022-II220311, Development of Goal-Oriented Reinforcement Learning Techniques for Contact-Rich Robotic Manipulation of Everyday Objects (31\%), No. RS-2024-00457882, AI Research Hub Project, No. RS-2019-II190079, Artificial Intelligence Graduate School Program (Korea University), the IITP (Institute of Information $\&$ Communications Technology Planning $\&$ Evaluation)-ITRC (Information Technology Research Center) grant funded by the Korea government (Ministry of Science and ICT) (IITP-2025-RS-2024-00436857) (31\%), the NRF (RS-2024-00451162) funded by the Ministry of Science and ICT, Korea, BK21 Four project of the National Research Foundation of Korea, the National Research Foundation of Korea (NRF) grant funded by the Korea government (MSIT) (RS-2025-00560367), the IITP under the Artificial Intelligence Star Fellowship support program to nurture the best talents (IITP-2025-RS-2025-02304828) (32\%) grant funded by the Korea government (MSIT), and KOREA HYDRO \& NUCLEAR POWER CO., LTD (No. 2024-Tech-09).

%% file: sec/X_suppl.tex
\clearpage
\setcounter{page}{1}
\maketitlesupplementary

\section{Related Works}
\label{related works}
\textbf{Vision-language pretraining (VLP)}
The training paradigm introduced by ALBEF \citep{li2021align}, which jointly optimizes the ITC, ITM, and MLM objectives, has served as the foundation for many subsequent VLP frameworks \citep{li2022blip, li2023blip, yang2022vision, zeng2021multi, bi2023vl, jian2023bootstrapping, jiang2023bus, byun2022grit, byun2024mafa, huang2024noise}. Given the high sensitivity of both ITC and ITM objectives to the difficulty of negative samples, several studies have proposed strategies to enhance vision-language pretraining by leveraging hard negative sampling. For example, ALBEF \citep{li2021align} computed the ITM loss by sampling in-batch hard negatives based on contrastive similarity scores computed within the current mini-batch. DiHT \citep{radenovic2023filtering} proposed an importance sampling approach to upweight harder negatives based on their similarity to the anchor. GRIT-VLP \citep{byun2022grit} enhances hard negative sampling by introducing the Grouped Mini-batch Sampling (GRIT) strategy, which constructs mini-batches composed of the most semantically similar image-text pairs retrieved from a large candidate pool $M$. This increases the chance of including informative hard negatives within each batch for both ITC and ITM losses. 

\textbf{False negatives in VLP}
While several prior works have proposed strategies to mitigate the impact of false negatives in the vision domain \citep{robinson2020contrastive, chuang2020debiased, chen2021incremental, huynh2022boosting, wu2020conditional}, the increased risk of false negatives introduced by hard negative sampling remains relatively underexplored in the context of vision-language pretraining. \cite{jiang2023vision} proposed Similarity-Regulated Contrastive Learning (SRCL), which adjusts the contrastive loss by weighting negative samples according to their cross-modal similarity to the anchor, where the similarity is initially estimated using a pretrained model and progressively refined during training. By assigning lower weights to semantically similar negatives, SRCL mitigates the over-penalization of false negatives during contrastive learning. More recently, \cite{byun2024mafa, bulat2024fff} demonstrated that converting false negatives into positives using a strong pretrained model can improve the performance on downstream tasks. These findings highlight a fundamental trade-off between hard and false negatives, emphasizing its significant impact on the learned representations. However, such methods rely heavily on pretrained models and fixed heuristic thresholds (e.g., ITM score cutoffs) to identify false negatives, which may limit their robustness and generalizability across diverse datasets and training conditions. In contrast, we propose a learning-based approach that adaptively balances the trade-off between hard and false negatives throughout the training process, without relying on fixed heuristics or external pretrained models. \cite{mei2025geomm} proposed a novel geodesic distance metric for multi-modal contrastive learning, designed to more accurately capture the underlying data manifold and thereby better distinguish positive and negative samples. Although this approach is conceptually compatible with FALCON’s hard negative scheduling mechanism, we were unable to evaluate its integration due to the absence of an official code release.

\textbf{Learning to optimize (L2O)} L2O is a research paradigm in machine learning that aims to automatically learn optimization algorithms from data, rather than relying on handcrafted update rules.  Early works in L2O typically adopted a meta-learning framework, where an optimizer is parameterized (e.g., via neural networks) and trained across a collection of optimization tasks \citep{chen2022learning, yang2023m, zheng2022symbolic}. In this framework, a meta-training set composed of multiple task-specific training and validation dataset pairs is used to guide the optimizer to generalize across tasks. Based on the meta-training set, L2O methods learn parameter update rules that minimize validation loss, either through supervised learning \citep{andrychowicz2016learning, wichrowska2017learned} or reinforcement learning \citep{bello2017neural, li2017learning}. Recent advances have begun to challenge these assumptions by introducing optimization policies that must learn and adapt in the absence of a predefined meta-training set \citep{kim2024adaptive}. In this paper, we propose an online optimization approach that constructs mini-batches to balance the tradeoff between hard and false negatives, without relying on any meta-training dataset.

\section{Experimental Details}
\label{experimental details}

\subsection{Experimental Setup}
Unless otherwise specified, all experiments follow the training protocols established in \citep{li2021align, byun2022grit, byun2024mafa}. For all retrieval tasks (COCO IRTR~\citep{lin2014microsoft}, Flickr IRTR~\citep{plummer2015flickr30k}), we evaluate the pretrained models directly without any task-specific fine-tuning. For NLVR2~\citep{suhr2018corpus}, we follow the protocol established in ALBEF~\citep{li2021align}, performing an additional pretraining stage on the COCO dataset to adapt the model for reasoning over paired images, followed by fine-tuning on the NLVR2 dataset for 10 epochs. For the VQA task~\citep{antol2015vqa}, we fine-tune the pretrained model for 8 epochs using both the training and validation splits of the COCO and Visual Genome datasets~\citep{krishna2017visual}, following standard practice in prior work~\citep{li2021align, byun2022grit, byun2024mafa}. To compute the ITC loss, we employed computationally efficient soft pseudo targets \citep{byun2022grit, byun2024mafa} instead of the pseudo targets generated by a momentum model \citep{li2021align} for computational efficiency. For the ITM loss, the negative sample with the highest similarity to the anchor within each mini-batch is selected as the negative \citep{byun2022grit, byun2024mafa}. For the IRTR task \citep{frome2013devise}, evaluation was performed on the MSCOCO 5K test set. Model training was primarily conducted on a machine equipped with four NVIDIA RTX 4090 GPUs.

\begin{algorithm}[t!]
\caption{Compose Mini-batch Index Set}
\label{alg:mini-batch construction}
\textbf{Input}: Similarity matrix $\mathbf{S}$, unselected index set $\mathcal{U}$, batch size $B$, scheduler $\pi_\phi$
\begin{algorithmic}[1]
\STATE Initialize mini-batch index set $\mathcal{I}=\{\}$
\STATE Select quantiles and normalize $\mathbf{S}$ to get $\widehat{\mathbf{S}}$
\STATE $q\sim\pi_\phi(\cdot\,\vert\, \widehat{\mathbf{S}})$
\STATE $i=\text{Uniform}( \mathcal{U})$
\STATE $\mathcal{I}\gets\mathcal{I}\cup\{i\}$
\STATE $\mathcal{U}\gets\mathcal{U}\,\backslash\,\{i\}$
\FOR{$B-1$}
\STATE $i \gets$ index of $q_{i}$-quantile of $\{ \mathbf{S}_{i, j} \,\vert\, j \in \mathcal{U} \}$
\STATE $\mathcal{I}\gets\mathcal{I}\cup\{i\}$
\STATE $\mathcal{U}\gets\mathcal{U}\,\backslash\,\{i\}$
\ENDFOR
\RETURN $\mathcal{I}$
\end{algorithmic}
% \vspace{-0.1cm}
\end{algorithm}

\subsection{Algorithms}
\label{algorithms}
This section presents the pseudo-code for mini-batch construction and the overall training loop of FALCON. Algorithm~\ref{alg:mini-batch construction} outlines the procedure for constructing a mini-batch index set $\mathcal{I}$. The process begins by sampling an initial anchor from the pool of unselected index set $\mathcal{U}$, followed by selecting the remaining $B-1$ indices based on quantile values $q$ drawn from the scheduler $\pi_\phi$. Algorithm~\ref{alg:vlp with sampler (search space)} describes the overall vision-language pretraining loop with a search space of size $|M|$. The image-text similarity matrix $\mathbf{S}$ is computed from the \texttt{[CLS]} embeddings in the current queue. Subsequently, a mini-batch is constructed using Algorithm~\ref{alg:mini-batch construction}. The vision-language model parameters $\theta$ are then updated via gradient descent, while the scheduler parameters $\phi$ are updated through gradient ascent.

\begin{algorithm}[t!]
\caption{VLP with Mini-batch Scheduler (for $i$-th search space $M$)}
\label{alg:vlp with sampler (search space)}
\textbf{Input}: VLP parameter $\theta$, scheduler parameter $\phi$, Vision dataset $\mathcal{V}$, Text dataset $\mathcal{T}$, learning rate $\eta, \gamma$

\begin{algorithmic}[1] %[1] enables line numbers
\STATE Compute pairwise similarity matrix $\mathbf{S}$ between $\mathcal{V}, \mathcal{T}$ in search space $M$
\STATE Initialize unselected index set $\mathcal{U}=\{0,\ldots,\vert M\vert-1\}$
\FOR{gradient step $k\in\{0,1,\ldots,\lfloor \vert M\vert/B\rfloor-1\}$}
\STATE Get mini-batch index $\mathcal{I}$ with Algorithm \ref{alg:mini-batch construction}
\STATE Construct mini-batch $V,T$ as samples at indices $\mathcal{I}+i\cdot\vert M\vert$ from $\mathcal{V},\mathcal{T}$
\STATE $\theta_{k+1}=\theta_k-\eta\cdot\nabla_{\theta_k}\mathcal{L}_\text{VLP}(V,T;\theta_k)$
\STATE $\Delta_k=\mathcal{L}_{\text{MLM}}(\theta_k)-\mathcal{L}_{\text{MLM}}(\theta_{k+1})$
\STATE $\phi_{k+1}=\phi_k+\gamma\cdot\Delta_k\nabla_{\phi_k}\log\pi_{\phi_k}(q\,\vert\, \widehat{\mathbf{S}})$
\ENDFOR
\end{algorithmic}
% \vspace{-0.1cm}
\end{algorithm}

\subsection{Implementation Details of the Mini-batch Construction Process}
\label{appd:details_minibatch}
At the beginning of vision-language pretraining, the queue does not yet contain a sufficient number of cached $\texttt{[CLS]}$ embeddings to construct the similarity matrix $\mathbf{S}$. Accordingly, we follow GRIT-VLP~\citep{byun2022grit, byun2024mafa} and adopt a standard uniform mini-batch sampling procedure during the first epoch, without training or applying the scheduler $\pi_\phi$. From the second epoch onward, cached image and text embeddings from the previous epoch are used to compute similarity matrices, enabling the scheduler to guide mini-batch construction. This design introduces a natural warm-start effect, providing a stable and efficient initialization for the learning-based mini-batch sampling scheduler $\pi_\phi$. As training progresses, the cached embeddings are updated epoch by epoch to reflect the current state of the VLP model, enabling the scheduler to make decisions that are aligned with the evolving structure of the embedding space.

To prevent overfitting, we apply instance-level scheduling at the end of each epoch to ensure that the search space does not consist of fixed instances throughout training \citep{byun2022grit}. This shuffling improves generalization by exposing the scheduler to a more diverse and representative set of training instances over time.

\begin{table}[t!]
    \centering
    \caption{Hyperparameter settings used for FALCON}
    \resizebox{\linewidth}{!}{
    \begin{tabular}{lr}
        \toprule
        \textbf{Hyperparameter} & \textbf{Setting}\\
        \midrule
         Image Resolution & 256  \\
         Embedding Dimension & 256 \\
         Batch Size $B$ & 96  \\
         Masking Probability & 0.5 \\
         Search Space size $\vert M\vert$ & \{2400, 5664, 28320\} \\
         Pretraining Epochs & 20  \\
         Optimizer  & AdamW($\beta=[0.9,0.999],\lambda=0.02$) \\
         learning rate $\gamma$  & scheduled \\
        \midrule
         $m$ for Subsampling & 100  \\
         Hidden Layer Dimension & 256 \\
         \# Residual Block & 2 \\
         Optimizer  & AdamW($\beta=[0.9,0.999],\lambda=0.01$) \\
         learning rate $\eta$ & 1e-4 \\
        \bottomrule
    \end{tabular}
    }
    \label{tab:hyperparams}
\end{table}
\subsection{Implementation Details of Baseline Methods}
\label{appd:details_baseline}

\begin{table*}[t!]
\centering
\caption{Retrieval performance of two baseline models (GRIT-VLP, MAFA) pretrained on the MSCOCO dataset under various hyperparameter configurations. All other settings are fixed.}
% \resizebox{\linewidth}{!}{
\begin{tabular}{l c 
                S[table-format=2.2] S[table-format=2.2]
                S[table-format=2.2] S[table-format=2.2]
                S[table-format=2.2] S[table-format=2.2]}
\toprule
\multirow{2}{*}{\textbf{Component}} & \multirow{2}{*}{\textbf{Setting}}
  & \multicolumn{3}{c}{\textbf{Text Retrieval}} 
  & \multicolumn{3}{c}{\textbf{Image Retrieval}} \\
  \cmidrule(lr){3-5} \cmidrule(lr){6-8}
  & & {R@1} & {R@5} & {R@10} & {R@1} & {R@5} & {R@10} \\
\midrule
\multirow{2}{*}{GRIT-VLP} 
  & $\vert M\vert=1920$     & 60.60 & 83.52 & 89.14 & 44.61 & 69.54 & 77.67 \\
  & $\vert M\vert=4800$     & 55.08 & 78.10 & 84.60 & 39.06 & 62.57 & 71.67 \\
\midrule
\multirow{3}{*}{MAFA}
  & $\vert M\vert=1920,\tau=0.98$ & 60.96 & 83.24 & 89.62 & 44.77 & 69.49 & 77.96 \\
  & $\vert M\vert=4800,\tau=0.98$ & 54.86 & 77.04 & 84.36 & 39.57 & 63.13 & 72.20 \\
  & $\vert M\vert=1920,\tau=0.80$ & 40.62 & 68.04 & 78.20 & 33.10 & 57.75 & 67.63 \\
\bottomrule
\end{tabular}%
% }
\label{tab:baseline_tuning}
\end{table*}

\begin{figure*}[t!]
    \centering
    \includegraphics[width=0.8\linewidth]{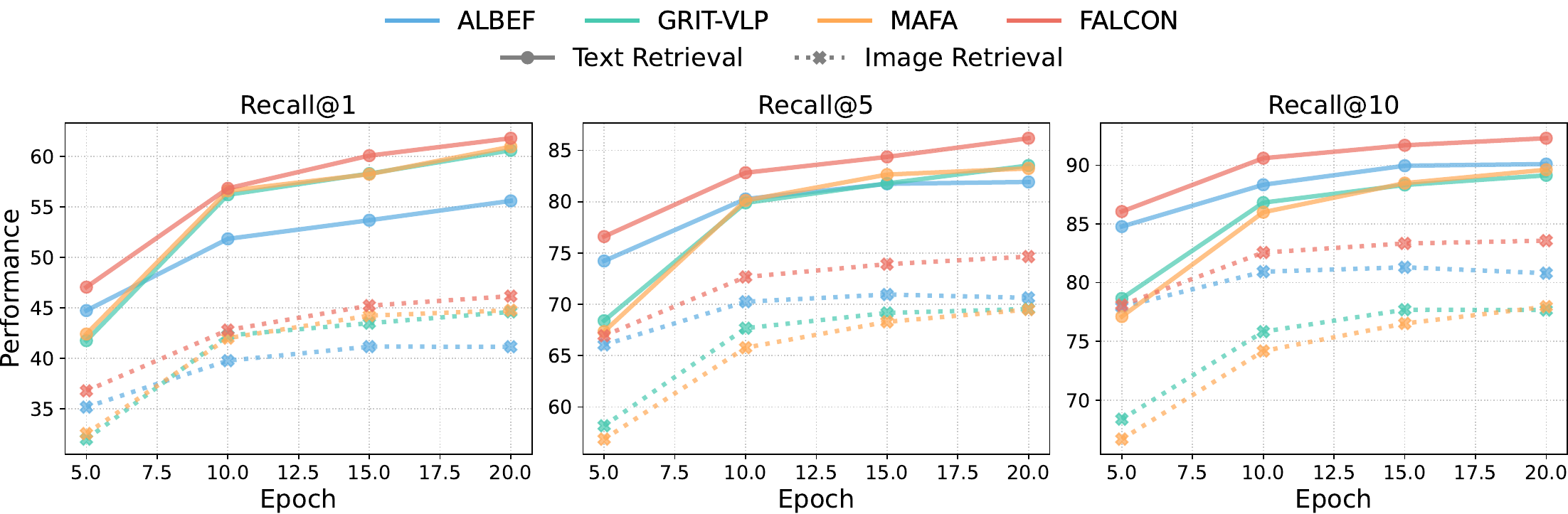}
    \caption{Performance comparison of VLP models across vision-language pretraining epochs on IRTR task. Recall@K (K = 1, 5, 10) is reported separately for text-to-image (solid lines) and image-to-text (dotted lines) retrieval.}
    \label{figure:mscoco_over_training}
\end{figure*}
For ALBEF~\citep{li2021align}, GRIT-VLP~\citep{byun2022grit}, MAFA~\citep{byun2024mafa}, and BLIP-2~\citep{li2023blip}, we conduct experiments using the official codebases released by the original authors.

For methods without publicly available implementations (DiHT~\citep{radenovic2023filtering} and SRCL~\citep{jiang2023vision}), we implemented the loss functions described in the respective papers using the ALBEF codebase as a foundation.

For quantile-based heuristic baselines, we adopt the mini-batch grouping procedure of GRIT-VLP, modifying the default quantile $q=1.0$ according to each heuristic strategy. These include fixed quantile settings, such as $q=0.5$ and $q=0.0$, as well as dynamic schedules in which the quantile is progressively increased (hardening) or decreased (softening) over the course of training.

For experiments involving SigLIP-2~\citep{tschannen2025siglip}, we initialized our models using the official pretrained checkpoint, as the training code has not been publicly released. We continued pretraining on the MSCOCO dataset using the hyperparameters reported in the original paper for five epochs. To ensure stable training, we excluded the SILC/TIPS loss, which we empirically found to cause instability during continued pretraining. Additionally, we replaced the original sigmoid-based contrastive loss with a softmax contrastive loss to more effectively exploit the benefits of hard negative batching.

\subsection{Hyperparameter Settings}
\label{hyperparameter settings}
We use the same backbone architecture and data augmentation strategy as ALBEF. The detailed hyperparameter settings are summarized in Table~\ref{tab:hyperparams}. For all remaining configurations, we follow the settings used in GRIT-VLP.

\subsection{Pretraining dataset size}
\label{Pretraining dataset size}
Table~\ref{tab:pretraining dataset size} shows the statistics of the pretraining dataset we used. For Conceptual Captions dataset, we used the preprocessed version provided by the original authors of BLIP~\citep{li2022blip}.%\footnote{\href{https://github.com/salesforce/BLIP?tab=readme-ov-file\#pre-training-datasets-download}{https://github.com/salesforce/BLIP?tab=readme-ov-file\#pre-training-datasets-download}}
\begin{table}[h!]
    \centering
    \caption{Statistics of the pretraining dataset}
    \begin{tabular}{l ccc}
        \toprule
         & \textbf{COCO} & \textbf{VG} & \textbf{CC + SBU}\\
        \midrule
        image & 113K & 100K & 3.63M \\
        text & 567K & 769K & 3.63M\\
        \bottomrule
    \end{tabular}
    \label{tab:pretraining dataset size}
\end{table}

\begin{table*}[t!]
    \centering
    \caption{Performance comparison of baseline models pretrained on the Conceptual Captions dataset \citep{sharma2018conceptual}, both with (left) and without (right) refinement using the BLIP captioner, in addition to the MSCOCO dataset.}
    \resizebox{\linewidth}{!}{
    \begin{tabular}{l|ccc ccc ccc ccc}
        \toprule
        \multirow{3}{*}{\textbf{Method}} & \multicolumn{6}{c}{\textbf{Clean (1.1M pretrain dataset)}} & \multicolumn{6}{c}{\textbf{Noisy (1.1M pretrain dataset)}}\\
        & \multicolumn{3}{c}{Text Retrieval} & \multicolumn{3}{c}{Image Retrieval} & \multicolumn{3}{c}{Text Retrieval} & \multicolumn{3}{c}{Image Retrieval}\\
        & R@1 & R@5 & R@10 & R@1 & R@5 & R@10 & R@1 & R@5 & R@10 & R@1 & R@5 & R@10 \\
        \midrule
        ALBEF          & 66.38 & 88.52 & 93.98 & 50.77 & 77.59 & 86.00 &63.92 &87.48 &93.20 &48.53 &75.96 &84.99 \\
        GRIT-VLP       & 66.42 & 86.62 & 91.90 & 48.90 & 73.21 & 80.60 &\textbf{67.92} &\textbf{88.66} &93.26 &50.00 &75.30 &82.65\\
        MAFA           & 66.36 & 88.60 & 92.90 & 50.70 & 74.83 & 83.66 & 65.94 & 85.58 & 90.52 & 49.39 & 72.42 & 79.33 \\
        \midrule
        FALCON         & \textbf{67.34} & \textbf{89.06} & \textbf{94.26} & \textbf{51.81} & \textbf{78.78} & \textbf{86.67} &66.00 &87.90 &\textbf{93.98} &\textbf{50.47} &\textbf{77.52} &\textbf{86.02} \\
        \bottomrule
    \end{tabular}
    }
    \label{table:blip-captioned}
    \vspace{-0.3cm}
\end{table*}

\begin{table}[t!]
\centering
\caption{Zero-shot MSCOCO performance comparison of models pretrained on the subset of DataComp dataset.}.
\resizebox{\linewidth}{!}{
\begin{tabular}{l 
                S[table-format=2.2] S[table-format=2.2]
                S[table-format=2.2] S[table-format=1.2]
                S[table-format=2.2] S[table-format=2.2]}
\toprule
\multirow{2}{*}{\textbf{Method}} 
  & \multicolumn{3}{c}{\textbf{Text Retrieval}} 
  & \multicolumn{3}{c}{\textbf{Image Retrieval}} \\
  \cmidrule(lr){2-4} \cmidrule(lr){5-7}
  & {R@1} & {R@5} & {R@10} & {R@1} & {R@5} & {R@10} \\
\midrule
  GRIT-VLP & 11.00 & 27.94 & 40.14 & 7.82 & 22.15 & 32.62 \\
  MAFA & 11.24 & 28.82 & 40.10 & 7.67 & 22.05 & 31.81 \\
  \midrule
  FALCON & \textbf{12.88} & \textbf{32.42} & \textbf{44.30} & \textbf{8.76} & \textbf{24.24} & \textbf{34.81} \\
\bottomrule
\end{tabular}
}
\label{tab:datacomp_retrieval}
\end{table}
\section{Additional Experiment results}
\label{appd:addtional_exp}

\subsection{Hyperparameter Sweeping in Baselines}
\label{appd:baseline sweep}
To ensure fair and competitive baselines, we sweep the search space size $\vert M \vert$ for both methods on the MSCOCO dataset and report in Table~\ref{tab:baseline_tuning}. For MAFA, we additionally sweep the similarity threshold $\tau$, which determines whether a given image-text pair is classified as a missed-positive (i.e., filtered out from negatives).

\subsection{Comparison with Heuristic Negative Mining Methods Across Training Epochs}
\label{appd:compare_epoch}
We visualize the learning curve of FALCON against baseline methods over the full course of epochs on the image-text retrieval (IRTR) downstream task in Figure~\ref{figure:mscoco_over_training}. All models are pretrained on the MSCOCO dataset. Throughout training, FALCON consistently outperforms all heuristic baselines across all epochs and recall metrics, highlighting its effectiveness in adaptively selecting negative samples with appropriate hardness during VLP mini-batch construction.

\begin{table}[t!]
\centering
\caption{Zero-shot 38 downstream tasks performance comparison of models pretrained on the subset of DataComp dataset.}
\label{tab:datacomp-38}
\resizebox{0.9\linewidth}{!}{
\begin{tabular}{l S[table-format=2.1] S[table-format=2.1] S[table-format=2.1]}
\toprule
\textbf{Dataset} & \textbf{GRIT-VLP} & \textbf{MAFA} & \textbf{FALCON} \\
\midrule
Caltech-101 & 28.9 & 33.9 & 35.3 \\
CIFAR-10 & 50.6 & 71.2 & 69.0 \\
CIFAR-100 & 10.1 & 23.6 & 25.4 \\
CLEVR Counts & 13.4 & 12.6 & 13.5 \\
CLEVR Distance & 24.6 & 22.2 & 20.9 \\
Country211 & 1.5 & 2.0 & 2.4 \\
Describable Textures & 11.3 & 17.2 & 15.4 \\
EuroSAT & 21.0 & 16.4 & 22.8 \\
FGVC Aircraft & 1.6 & 1.2 & 1.6 \\
Food-101 & 4.9 & 8.2 & 3.9 \\
GTSRB & 3.7 & 5.8 & 4.0 \\
ImageNet 1k & 7.4 & 11.9 & 14.5 \\
ImageNet Sketch & 2.7 & 4.6 & 5.9 \\
ImageNet v2 & 6.4 & 9.9 & 12.1 \\
ImageNet-A & 3.3 & 5.0 & 5.8 \\
ImageNet-O & 13.5 & 18.0 & 22.0 \\
ImageNet-R & 5.7 & 7.1 & 8.1 \\
KITTI Vehicle Distance & 11.3 & 10.0 & 10.9 \\
MNIST & 7.4 & 9.1 & 10.5 \\
ObjectNet & 2.5 & 3.4 & 3.5 \\
Oxford Flowers-102 & 2.7 & 1.7 & 1.7 \\
Oxford-IIIT Pet & 7.4 & 10.0 & 7.7 \\
Pascal VOC 2007 & 33.0 & 35.4 & 39.4 \\
PatchCamelyon & 51.3 & 52.0 & 50.8 \\
Rendered SST2 & 50.0 & 49.9 & 50.1 \\
RESISC45 & 16.2 & 13.7 & 15.3 \\
Stanford Cars & 2.3 & 3.6 & 3.3 \\
STL-10 & 75.8 & 81.9 & 83.3 \\
SUN397 & 9.3 & 11.2 & 13.2 \\
SVHN & 7.8 & 7.3 & 7.3 \\
Flickr & 9.9 & 12.7 & 14.9 \\
MSCOCO & 5.3 & 7.1 & 7.7 \\
WinoGAViL & 37.7 & 37.4 & 40.2 \\
iWildCam & 0.5 & 0.7 & 1.6 \\
Camelyon17 & 51.4 & 50.9 & 51.7 \\
FMoW & 0.0 & 1.9 & 0.0 \\
Dollar Street & 33.9 & 34.9 & 38.8 \\
GeoDE & 33.5 & 48.0 & 45.6 \\
\midrule
\textbf{Average} & 18.1 & 20.7 & \textbf{21.3} \\
\bottomrule
\end{tabular}
}
\end{table}

\begin{table*}[t!]
    \centering
    \caption{Performance Comparison of FALCON with baselines under the BLIP-2 framework.}
    \resizebox{\linewidth}{!}{
    \begin{tabular}{c | ccc ccc cc | c c c cc}
        \toprule
        \multirow{3}{*}{\textbf{Method}} & \multicolumn{8}{c|}{\textbf{Stage-1}} & \multicolumn{5}{c}{\textbf{Stage-2}} \\
        & \multicolumn{3}{c}{\textbf{COCO Text Retrieval}} & \multicolumn{3}{c}{\textbf{COCO Image Retrieval}} & \multicolumn{2}{c|}{\textbf{Flickr R@1}} & \textbf{VQA2} & \textbf{OKVQA} & \textbf{GQA} & \multicolumn{2}{c}{\textbf{Captioning}}\\
        & R@1 & R@5 & R@10 & R1 & R@5 & R@10 & TR & IR & val & test & test-dev & CIDEr & SPICE\\
        \midrule
        BLIP-2          & 75.22 & 93.00 & 96.50 & 57.98 & 82.08 & 88.78 & 90.10 & 77.48 & 42.46 & 17.94 & 28.87 & \textbf{107.2} & \textbf{19.5} \\
        + GRIT-VLP   & 73.90 & 93.10 & 96.52 & 57.47 & 80.50 & 87.56 & 90.40 & 77.28 & 39.91 & 15.92 & 27.75 & 105.9 & 19.4 \\
        + MAFA   & 74.21 & 93.00 & 96.61 & 57.94 & 81.12 & 88.44 & 90.30 & 77.32 & 41.12 & 18.54 & 28.93 & 106.1 & 19.3 \\
        + FALCON & \textbf{75.56} & \textbf{93.50} & \textbf{96.90} & \textbf{58.52} & \textbf{82.39} & \textbf{88.98} & \textbf{90.90} & \textbf{77.72} & \textbf{42.67} & \textbf{20.96} & \textbf{29.29} & 106.0 & 19.4 \\
        \bottomrule
    \end{tabular}
    }
    \label{table:blip-detail}
\end{table*}

\begin{figure*}[t!]
    \centering      
    \includegraphics[width=1.0\linewidth]{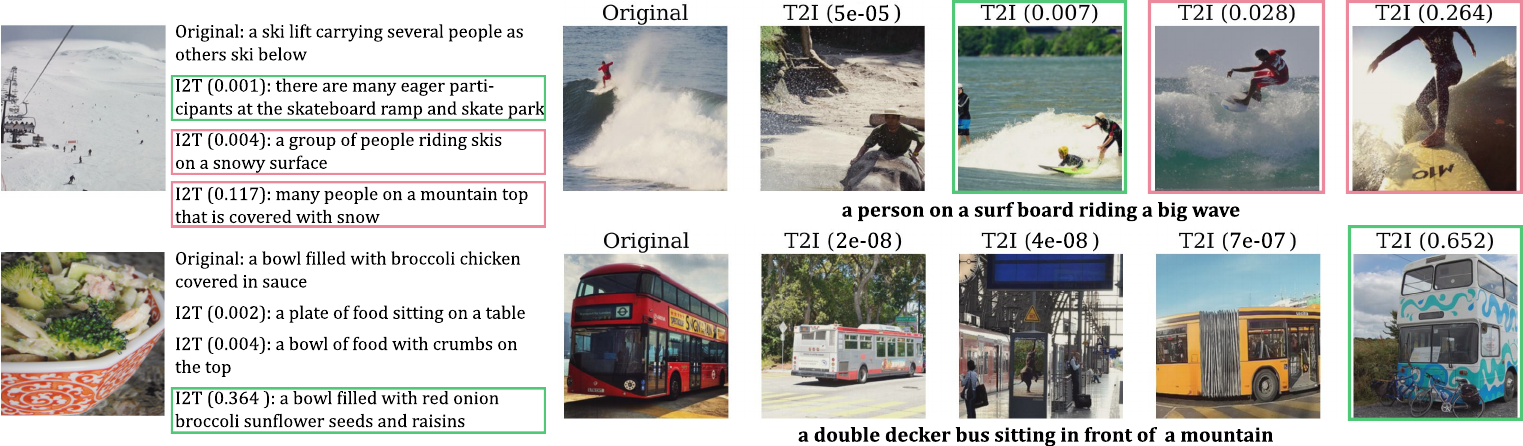}
    \caption{Image-to-Text \textbf{(Left)} and Text-to-Image \textbf{(Right)} examples of negative sampling under FALCON’s quantile-based scheduling strategy. Negative candidates are drawn from similarity score quantiles [0.8, 0.9, 1.0] for I2T and [0.5, 0.8, 0.9, 1.0] for T2I. The negative sample selected by FALCON is highlighted in green and the genuine false negative sample is highlighted in red.}
    \label{app:falcon_negative}
    \vspace{-0.3cm}
\end{figure*}
\subsection{Comparison on BLIP-Captioned and Noisy Datasets} 
To further evaluate the robustness of FALCON in mitigating false negatives, we conducted experiments on web-crawled image-text pairs from the Conceptual Captions dataset 
\label{appd:blip_captioned}
\citep{sharma2018conceptual}, with and without refinement using the BLIP captioner \citep{li2022blip}, in addition to the MSCOCO dataset. This evaluation assess whether FALCON can leverage high-quality captions generated by BLIP to improve performance on noisy, web-crawled data. As shown in Table~\ref{table:blip-captioned} (left), FALCON significantly outperforms heuristic baselines and demonstrates further performance gains as caption quality increases, compared to the results in Table~\ref{table:supple-overall-baseline}. However, when the same web-crawled data is used without BLIP-based refinement, the performance gains become less pronounced, particularly in the text retrieval task (Table~\ref{table:blip-captioned} (right)). We attribute this to significant noise and semantic misalignment in the original captions, which can hinder accurate estimation of the tradeoff between hard and false negatives, thereby making the learning process more complex.

\subsection{Comparison on DataComp dataset}
\label{appd:datacomp}

To demonstrate the generality and robustness of FALCON, we pretrained FALCON and baselines on a 1M subset of 12.8B DataComp~\citep{gadre2023datacomp} filtered by CLIP similarity and English filtering. We evaluate zero-shot performance on the MSCOCO image-text retrieval benchmark, as well as on the full suite of 38 classification and retrieval tasks from the DataComp benchmark. The results are reported in Table~\ref{tab:datacomp_retrieval} and Table~\ref{tab:datacomp-38}, respectively. 

For comparison under a matched data budget, Table~\ref{table:blip-captioned} (left) presents results on the filtered Conceptual Captions dataset of approximately 1.1M pairs—comparable in scale to the 1M DataComp subset. Under this controlled setting, FALCON achieves an average improvement of 2.4\% over MAFA in Table~\ref{table:blip-captioned}, while this margin increases to 11.9\% in Table~\ref{tab:datacomp_retrieval}. 

We attribute this discrepancy to a key limitation of MAFA, its reliance on a fixed pretrained model (BLIP-129M) for false-negative detection. This reliance inherently couples its effectiveness to the domain and distributional characteristics of the data used during the pretraining of the filtering model. In our experiments, MAFA's performance degrades significantly when applied to DataComp dataset, which deviates significantly from the BLIP-129M model's original pretraining corpus. These findings underscore the need for adaptive methods like FALCON, which can dynamically identify false negatives based on the current training data, regardless of prior pretraining exposure.

\begin{table}[t]
\centering
\caption{Performance on downstream vision-and-language tasks.}
\resizebox{\linewidth}{!}{
\begin{tabular}{l|c|cccc}
\toprule
\multirow{2}{*}{Method} & \multirow{2}{*}{\#Images} & \multicolumn{2}{c}{VQA} & \multicolumn{2}{c}{NLVR$^2$} \\
 &  & test-dev & test-std & dev & test-P \\
\midrule
UNITER~\cite{chen2020uniter} & 4M & 72.70 & 72.91 & 77.18 & 77.85 \\
OSCAR~\cite{li2020oscar}     & 4M & 73.16 & 73.44 & 78.07 & 78.36 \\
VILLA~\cite{gan2020large}    & 4M & 73.59 & 73.67 & 78.39 & 79.30 \\
ViLT~\cite{kim2021vilt}      & 4M & 70.94 & --    & 75.24 & 76.21 \\
ALBEF~\cite{li2021align}     & 4M & 74.54 & 74.70 & 80.24 & 80.50 \\
TCL~\cite{yang2022vision}    & 4M & 74.90 & 74.92 & 80.54 & 81.33 \\
GRIT-VLP~\cite{byun2022grit} & 4M & 75.11 & 75.26 & 80.73 & 81.60 \\
MAFA~\cite{byun2024mafa}     & 4M & 75.55 & 75.75 & 82.52 & 82.08 \\ 
\midrule
FALCON                       & 4M & \textbf{75.62} & \textbf{75.78} & \textbf{82.61} & \textbf{82.28} \\
\bottomrule
\end{tabular}
}
\label{tab:vltasks}
\vspace{-0.2cm}
\end{table}

\section{Limitations and Future work}
\label{appd:limitations}
As discussed in Section~\ref{exp:compatibility_with_SIGLIP}, our findings suggest that for FALCON to be fully effective, the proxy signal used for cross-modal alignment should integrate information from both the vision and text encoders. A promising future direction is to develop learning-based strategies for scheduling the trade-off between hard and false negatives in vision–language pretraining that do not rely on such auxiliary objectives, thereby enabling broader applicability across contrastive learning paradigms~\citep{radford2021learning}.

Furthermore, the recent emergence of Large Vision-Language Models (LVLMs) has demonstrated strong performance across a wide range of multimodal tasks. These models are typically built upon VLP backbones that provide the core cross-modal representations~\citep{bai2025qwen2, li2024llava, wu2024deepseek}. We believe that continued improvements in these VLP backbones, along with advances in contrastive learning (e.g., false-negative-aware strategies like FALCON), will contribute to the future development and effectiveness of LVLMs.

\section{Additional Visualization of FALCON}
\clearpage
\label{appd:ours_viz}
\begin{figure*}[h]
    \centering
    \vspace{-0.3cm}
    \includegraphics[width=0.9\linewidth]{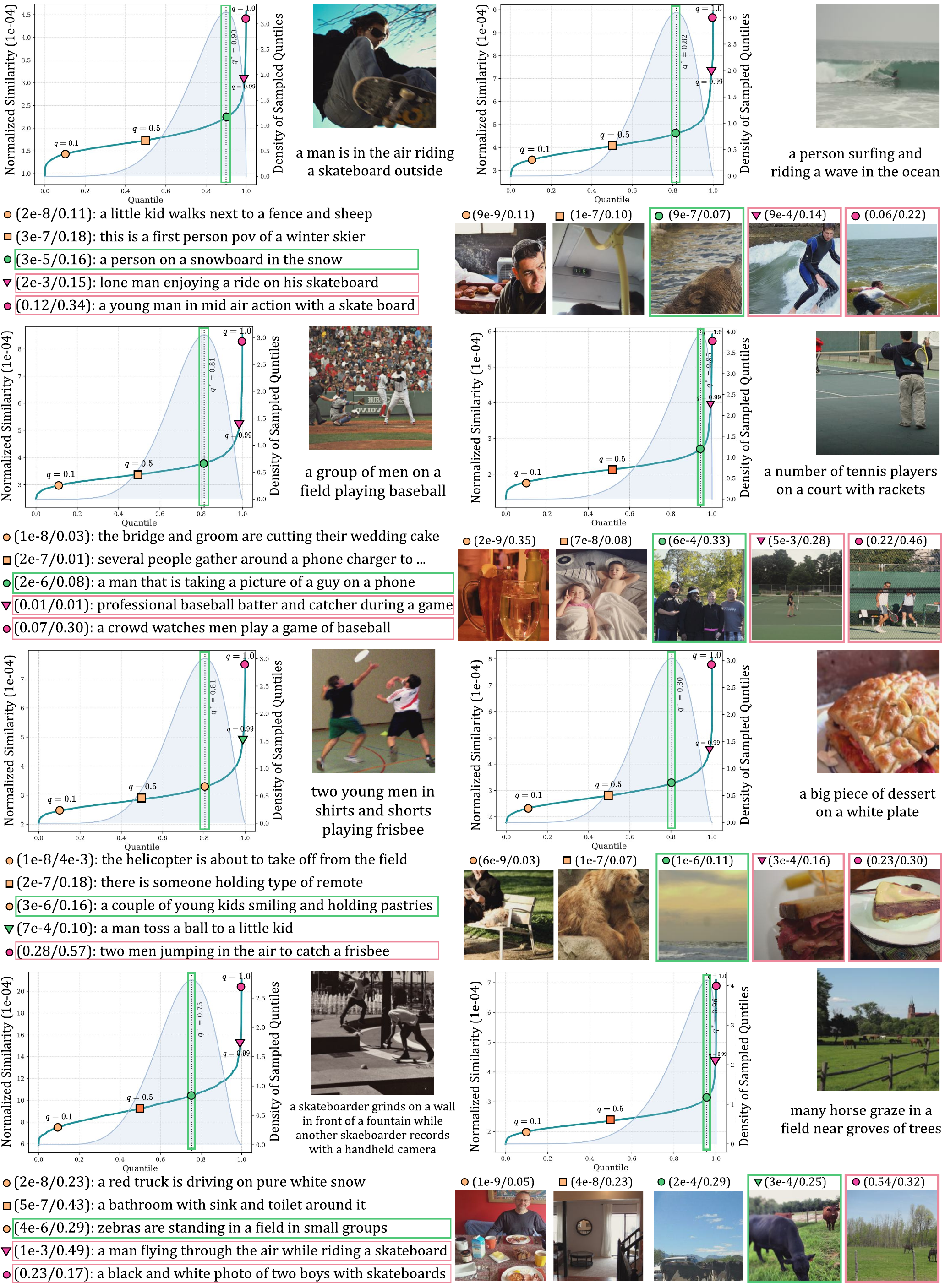}
    \caption{Additional anchor-specific negative sampling visualizations. We highlight the mode of scheduler distribution in green and genuine false negatives in red. Each negative is annotated with ``(one-way similarity / ITM score)" and its hardness is color-coded as in Figure~\ref{fig:relationship}.}
    \label{fig:enter-label}
\end{figure*}